\documentclass[a4paper, orivec, runningheads]{llncs}
\usepackage{floatrow}
\newfloatcommand{capbtabbox}{table}[][\FBwidth]
\usepackage{tablefootnote}
\usepackage{graphicx}
\usepackage{amsmath}
\usepackage{bm}
\usepackage{float}
\usepackage{url}
\usepackage{makeidx}  
\usepackage{graphicx}
\usepackage{subfigure}
\usepackage{amsmath}
\usepackage{amssymb}
\usepackage{verbatim}
\usepackage{algorithm}
\usepackage{algorithmic}
\usepackage{booktabs}
\usepackage{multirow}
\usepackage{pbox} 
\usepackage{cite}
\usepackage{wrapfig}
\usepackage{threeparttable}
\usepackage{appendix}
\usepackage[colorlinks,
            linkcolor=blue,
            anchorcolor=blue,
            citecolor=blue
            ]{hyperref}
\usepackage{txfonts}
\begin{document}
\title{Efficient Semi-Supervised Gross Target Volume of Nasopharyngeal Carcinoma Segmentation via Uncertainty Rectified Pyramid Consistency}
\titlerunning{Uncertainty Rectified Pyramid Consistency for NPC}
\authorrunning{Luo et al.} %
\author{Xiangde Luo\inst{1} \and Wenjun Liao \inst{2}\and Jieneng Chen\inst{3} \and Tao Song\inst{4} \and Yinan Chen\inst{4,6} \and Shichuan Zhang\inst{5} \and Nianyong Chen\inst{2} \and Guotai Wang\inst{1} \and Shaoting Zhang\inst{1, 4}}

\institute{
$^1$School of Mechanical and Electrical Engineering, University of Electronic Science and Technology of China, Chengdu, Chengdu, China\\
$^2$West China Hospital, Sichuan University, Chengdu, China\\
$^3$College of Electronics and Information Technology, Tongji University, Shanghai, China\\
$^4$SenseTime Research, Shanghai, China\\
$^5$Sichuan Cancer Hospital \& Institute\\
$^6$West China Hospital-SenseTime Joint Lab, West China Biomedical Big Data Center, Sichuan University
West China Hospital, Chengdu, China\\
\url{https://github.com/HiLab-git/SSL4MIS}\\
Corresponding author: \url{guotai.wang@uestc.edu.cn}\\
}
\maketitle

\begin{abstract}
    Gross Target Volume (GTV) segmentation plays an irreplaceable role in radiotherapy planning for Nasopharyngeal Carcinoma (NPC). Despite that Convolutional Neural Networks (CNN) have achieved good performance for this task, they rely on a large set of labeled images for training, which is expensive and time-consuming to acquire. In this paper, we propose a novel framework with Uncertainty Rectified Pyramid Consistency (URPC) regularization for semi-supervised NPC GTV segmentation. Concretely, we extend a backbone segmentation network to produce pyramid predictions at different scales. The pyramid predictions network (PPNet) is supervised by the ground truth of labeled images and a multi-scale consistency loss for unlabeled images, motivated by the fact that prediction at different scales for the same input should be similar and consistent. However, due to the different resolution of these predictions, encouraging them to be consistent at each pixel directly has low robustness and may lose some fine details. To address this problem, we further design a novel uncertainty rectifying module to enable the framework to gradually learn from meaningful and reliable consensual regions at different scales. Experimental results on a dataset with 258 NPC MR images  showed that with only 10\% or 20\% images labeled, our method largely improved the segmentation performance by leveraging the unlabeled images, and it also outperformed five state-of-the-art semi-supervised segmentation methods.  Moreover, when only 50\% images labeled, URPC achieved an  average Dice score of 82.74\% that was close to fully supervised learning.
\end{abstract}

\section{Introduction}
Nasopharyngeal Carcinoma (NPC) is one of the most common cancers in southern China, Southeast Asia, the Middle East, and North Africa~\cite{chen2016cancer}. The mainstream treatment strategy for NPC is radiotherapy, thus the accurate target delineation plays an irreplaceable role for precise and effective radiotherapy. However, manual nasopharyngeal tumor contouring is tedious and laborious, since both the nasopharynx gross tumor volume (GTVnx) and lymph node gross tumor volume (GTVnd) need to be accurately delineated~\cite{lin2019deep}. Recently, with a large amount of labeled data, deep learning has shown the potential for accurate GTV segmentation~\cite{lin2019deep}. However, collecting a large labeled dataset for network training is difficult, as both time and expertise are needed to produce accurate annotation. In contrast, collecting a large set of unlabeled data is easier, which inspired us to develop a semi-supervised approach for NPC GTV segmentation by leveraging unlabeled data. What's more, semi-supervised learning (SSL) can largely reduce the workload of annotators for the development of deep learning models.
\par Recently, SSL has been widely used for medical image computing to reduce the annotation efforts~\cite{li2020transformation,ma2020active,hang2020local,peng2020mutual}. Bai et al~\cite{bai2017semi} developed an iterative framework where in each iteration, pseudo labels for unannotated images are predicted by the network and refined by a Conditional Random Field (CRF), then the new pseudo labels are used to update the network. After that, the perturbation-based methods have achieved increasing attentions in semi-supervised learning~\cite{li2020transformation,chaitanya2019semi, bortsova2019semi, peng2020mutual}. These methods add small perturbations to unlabeled samples and enforce the consistency between the model's predictions on the original data and the perturbed data. Meanwhile, the mean teacher-based~\cite{tarvainen2017mean} self-ensembling methods~\cite{cui2019semi, yu2019uncertainty, wang2020double, hang2020local} were introduced for semi-supervised medical image segmentation. Following~\cite{cui2019semi}, some recent works~\cite{yu2019uncertainty, wang2020double, hang2020local} used uncertainty map to guide the student model to learn more stably. In~\cite{zhang2017deep,li2020shape,nie2018asdnet}, an adversarial training strategy was used as regularization for SSL, which aims to minimize the adversarial loss to encourage the prediction of unlabeled data is anatomical plausible. Luo et al.~\cite{luo2020semi} proposed a dual-task consistency framework for SSL by representing segmentation as a pixel-wise classification task and a level set regression task simultaneously, the difference between which was minimized during training. Despite their higher performance than learning from available labeled images only, existing methods are limited by high computational cost and complex training strategies in practice. For example, the co-training-based methods need to train several models at the same time~\cite{qiao2018deep}, and the uncertainty estimation-based frameworks need multiple forward passes~\cite{yu2019uncertainty}. Self-training-based approaches need to select and refine pseudo labels and update models' parameters in several rounds~\cite{bai2017semi}, which is time consuming.

\par In this work, we propose a novel efficient semi-supervised learning framework for the segmentation of GTVnx and GTVnd by further utilizing the unlabeled data. Our method leverages a network that gives a pyramid (i.e., multi-scale) prediction, and encourages the predictions at multiple scales to be consistent for a given input, which is a simple yet efficient idea for SSL. A standard supervised loss at multiple scales is used for learning from labeled images. For unlabeled images, we encourage the multi-scale predictions to be consistent, which serves as a regularization. Since the ground truth of unlabeled images is unknown, the model may produce some unreliable prediction or noise which may cause the model to collapse and lose details. To overcome these problems, some existing works~\cite{yu2019uncertainty,cao2020uncertainty} have leveraged model uncertainty to boost the stability of training and obtain better results. However, they typically estimate the uncertainty of each target prediction with Monte Carlo sampling~\cite{gal2016dropout}, which needs massive computational costs as it requires multiple forward passes to obtain the uncertainty in each iteration. Differently from these methods, we estimate the uncertainty via the prediction discrepancy among multi-scale predictions, which just needs a single forward pass. With the guidance of the estimated uncertainty, we automatically emphasize the reliable predictions (low uncertainty)  and weaken the unreliable ones (high uncertainty) when calculating the multi-scale consistency. Meanwhile, we introduce the uncertainty minimization~\cite{zheng2020rectifying} to reduce the prediction variance during training. Therefore, the proposed framework has a high efficiency for semi-supervised segmentation by taking advantage of the unlabeled images. Our method was extensively evaluated on a clinical Nasopharyngeal Carcinoma dataset. Results show our method largely improved the segmentation performance by leveraging the unlabeled images, and it outperformed five state-of-the-art semi-supervised segmentation methods. Moreover, when only half of the training images are labeled, URPC achieves a very close result compared with fully supervised learning (the mean of Dice was 82.74\% vs 83.51\%).

\begin{figure}[t]
	\centering
	\includegraphics[width=1.0\linewidth]{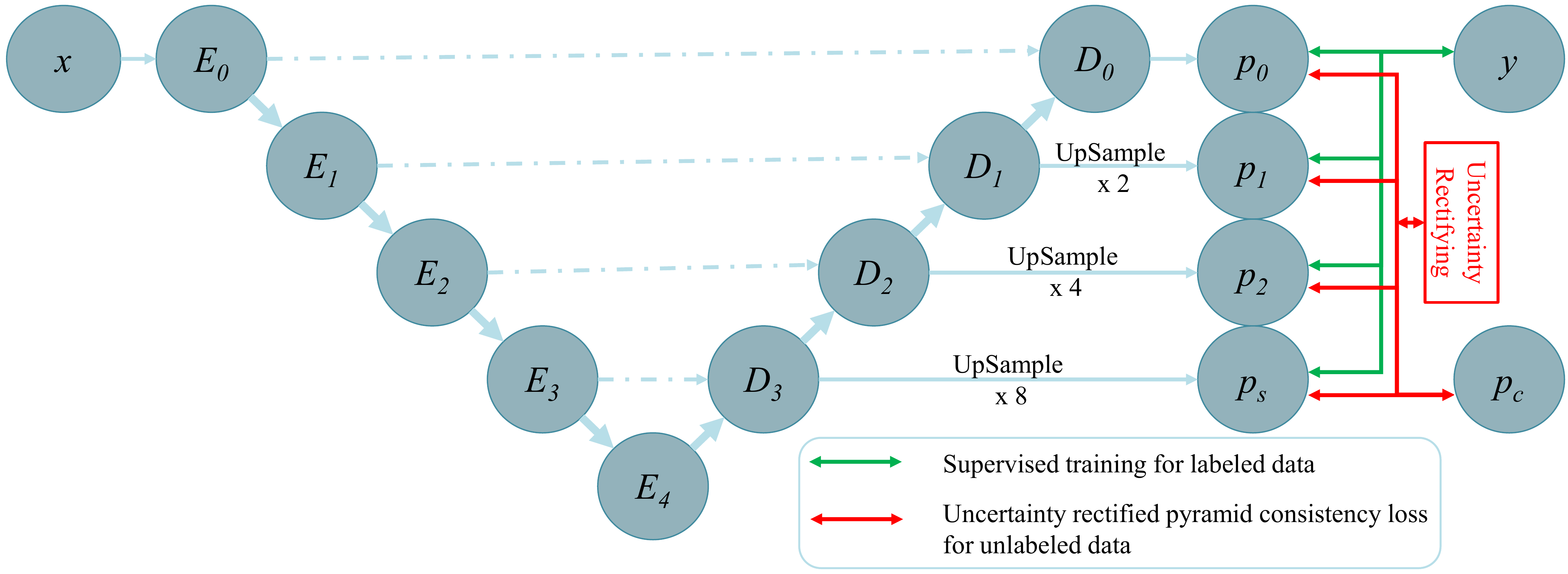}
	\caption{Overview of the proposed novel Uncertainty Rectified Pyramid Consistency framework, which consists of a pyramid prediction network (PPNet) and an uncertainty rectifying module. It is based on a backbone of 3D UNet~\cite{cciccek20163d}, where $Es$ and $Ds$ are the blocks in the encoder and decoder of 3D UNet respectively. $p_s$ is the prediction as scale $s$. The URPC is optimized by minimizing the supervised loss on the labeled data and the pyramid consistency loss on the unlabeled data. In addition, an uncertainty rectifying module is designed to reduce the impact of noise in the pyramid consistency and boost the stability of training.}
	\label{overview}
\end{figure}

\section{Methods}
The proposed URPC for semi-supervised segmentation is illustrated in Fig.~\ref{overview}. We add a pyramid prediction structure at the decoder of a backbone network and refer to it as PPNet. PPNet learns from the labeled data by minimizing a typical supervised segmentation loss directly. In addition, the PPNet is regularized by a multi-scale consistency between the pyramid predictions to deal with unlabeled data. The PPNet naturally leads to uncertainty estimation in a single forward pass by measuring the discrepancy between these predictions, and we propose to use this uncertainty to rectify the pyramid consistency considering the different spatial resolutions in the pyramid. To describe this work precisely, we firstly define some mathematical terms. Let ${D}_{l}$, ${D}_{u}$ and $f_\phi(x)$ be the labeled set,  unlabeled set and the PPNet's parameters, respectively. Let 
${D} = {D}_{l} \cup {D}_{u}$ be the whole provided dataset. We denote an unlabeled image as $x_i \in {D}_{u}$ and a labeled image pair as $(x_i, y_i) \in {D}_{l}$, where $y_i$ is ground truth. 
\subsection{Multi-Scale Prediction Network with Pyramid Consistency}To better exploit the prediction discrepancy of a single model at different scales, we firstly introduce the PPNet for the segmentation task, which can produce predictions with different scales. In this work, we employ 3D UNet~\cite{cciccek20163d} as a backbone and modify it to produce pyramid predictions by adding a prediction layer after each upsampling block in the decoder, where the prediction layer is implemented by $1 \times 1 \times 1$ convolution followed by a softmax layer. To introduce more perturbations in the network, a dropout layer and a feature-level noise addition layer are inserted before each of these prediction layers. For an input image $x_i$, PPNet $f_\phi(x)$ produces a set of multi-scale predictions $\left[p_0', p_1', ..., p_s', ..., p_{S-1}'\right]$, where the $p'_s$ is the prediction at scale $s$, and a smaller $s$ corresponds to a higher resolution in the decoder, as shown in Fig.~\ref{overview}. $S$ is the number of scales in the pyramid prediction. Then, we rescale these multi-scale predictions to the input size, and the corresponding results are denoted as $\left[p_0, p_1, ..., p_s, ..., p_{S-1}\right]$. For the labeled data, we use a supervised loss that is a combination of Dice and cross-entropy loss at multiple scales:
\begin{equation}\label{equ:sup_loss}
    \mathcal{L}_{sup} = \frac{1}{S}\sum_{s=0}^{S-1} \frac{\mathcal{L}_{dice}(p_s, y_i) + \mathcal{L}_{ce}(p_s, y_i)}{2}
\end{equation}where $y_i$, $\mathcal{L}_{dice}$, $\mathcal{L}_{ce}$ denote the ground truth of input $x_i$, the Dice loss and the cross entropy loss, respectively. 
\par To efficiently leverage unlabeled data, we introduce a regularization by encouraging the multi-scale predictions of PPNet to be consistent. Concretely, we design a pyramid consistency loss to minimize the discrepancy (i.e., variance) among the predictions at different scales. First, we denote the average prediction across these scales as:
\begin{equation}\label{equ:cons_pred}
    p_c =  \frac{1}{S}\sum_{s=0}^{S-1}p_s
\end{equation}Then, the pyramid consistency loss is defined as:
\begin{equation}\label{equ:loss_unsup}
    \mathcal{L}_{pyc} = \frac{1}{S}\sum_{s=0}^{S-1}\left\|p_s - p_c\right\|_{2}
\end{equation}where we encourage a minimized $L_2$ distance between the prediction at each scale and the average prediction. 

\subsection{Uncertainty Rectified Pyramid Consistency Loss}
As the pyramid prediction at a range of scales have different spatial resolutions, even they can be resampled to the same resolution as the input, the resampled results still have different spatial frequencies, i.e, the prediction at the lowest resolution captures the low-frequency component of the segmentation, and the prediction at the highest resolution obtains more high-frequency components. Directly imposing a voxel-level consistency among these predictions can be problematic due to the different frequencies, such as lost of fine details or model collapse. Inspired by existing works~\cite{yu2019uncertainty,cao2020uncertainty,wang2020uncertainty,wang2019aleatoric,zheng2020rectifying}, we introduce an uncertainty-aware method to address these problems. Unlike existing methods, our uncertainty estimation is a scale-level approach and only requires a single forward pass, which needs less computational cost and running time than exiting methods.
\subsubsection{Efficient Uncertainty Estimation based on Pyramid Predictions.}
As our PPNet obtains multiple predictions in a single forward pass, uncertainty estimation can be obtained efficiently by justing measuring their discrepancy without extra efforts. To be specific, we use the KL-divergence between the average prediction and the prediction at scale $s$ as the uncertainty measurement:
\begin{equation}\label{equ:unc}
    \mathcal{D}_{s} \approx \sum_{j=0}^{C} p_s^j\cdot \log\frac{p_s^j}{p_c^j}
\end{equation}where $p^j_s$ is the $j~th$ channel of $p_s$, and $C$ is the class (i.e., channel) number. The approximated uncertainty shows the difference between the $p_s$ and $p_c$. Note that for a given voxel in $\mathcal{D}_s$, a larger value indicates the prediction for that pixel at scale $s$ is far from the other scales, i.e., with high uncertainty. As result, we obtain a set of uncertainty maps $\mathcal{D}_0$, $\mathcal{D}_1$, ... $\mathcal{D}_{S-1}$, where $\mathcal{D}_s$  corresponds  to uncertainty of $p_s$. 
\subsubsection{Uncertainty Rectifying.}Based on the estimated uncertainty maps $\mathcal{D}_0$, $\mathcal{D}_1$, ... $\mathcal{D}_{S-1}$, we further extend the pyramid consistency $\mathcal{L}_{pyc}$ to emphasize reliable parts and ignore unreliable parts of the predictions for stable unsupervised training. Specifically, for unlabeled data, we use the estimated uncertainty to automatically select reliable voxels for loss calculation. The rectified pyramid consistency loss is formulated as:
\begin{equation}\label{equ:loss_rec}
    \mathcal{L}_{unsup} = \underbrace{\frac{1}{S} \frac{\sum_{s=0}^{S-1} \sum_{v} (p_s^v - p_c^v)^2 \cdot w_s^v }{\sum_{s=0}^{S-1}\sum_{v} w_s^v}}_{uncertainty~rectification}~+~\underbrace{\frac{1}{S} \sum_{s=0}^{S-1}|| \mathcal{D}_{s}||_2}_{uncertainty~minimization}
\end{equation}where $p^v_s$ and $\mathcal{D}^v_s$ are the corresponding prediction and uncertainty values for voxel v. The consistency loss consists of two terms: the first is an uncertainty rectification (UR) term and the second is uncertainty minimization (UM) term. For a more stable training, we follow the policy in~\cite{zheng2020rectifying} and we use a voxel- and scale-wise weight $w^v_s$ to automatically rectify the MSE loss rather than the threshold-based cut off approaches~\cite{yu2019uncertainty,cao2020uncertainty}, as the threshold is hard to determine. The weight for a voxel $v$ at scale $s$ is defined as: $w_s^v = e^{-\mathcal{D}_{s}^{v}}$ \unboldmath, it corresponds to voxel-wise exponential operation for -~$\mathcal{D}_s$. According to this definition, for a given voxel at scale $s$, a higher uncertainty leads to a lower weight automatically. In addition, to encourage the PPNet to produce more consistent predictions at different scales, we use the uncertainty minimization term as a constraint directly. With this uncertainty rectified consistency loss, the PPNet can learn more reliable knowledge, which can then reduce the overall uncertainty of the model and produce more consistent predictions.
\subsection{The Overall Loss Function}The proposed URPC framework learns from both labeled data and unlabeled data by minimizing the following combined objective function:
\begin{equation}\label{equ:total_loss}
    \mathcal{L}_{total} = \mathcal{L}_{sup} + \lambda \cdot \mathcal{L}_{unsup}
\end{equation}where $\mathcal{L}_{sup}$, $\mathcal{L}_{unsup}$ are defined in Eq.~\ref{equ:sup_loss} and Eq.~\ref{equ:loss_rec}, respectively. $\lambda$ is a widely-used time-dependent Gaussian warming up function~\cite{tarvainen2017mean,yu2019uncertainty}  to control the balance between the supervised loss and unsupervised consistency loss, which is defined as $\lambda(t) = w_{max}\cdot e^{(-5(1-\frac{t}{t_{max}})^2)}$, where $w_{max}$ means the final regularization weight, $t$ denotes the current training step and $t_{max}$ is the maximal training step.
\section{Experiments and Results}
\subsection{Dataset and Implementations} 
The NPC dataset used in this work was collected from the West China Hospital. A total number of 258 T1-weighted MRI images from 258 patients of NPC before radiotherapy were acquired on several 3T Siemens scanners. The mean resolution of the dataset was 1.23 mm×1.23 mm×1.10 mm, and the mean dimension was 176×286×245. The ground truth for GTVnx and GTVnd were obtained from manual segmentation by two experienced radiologists using ITK-SNAP~\cite{yushkevich2006user}. The dataset was randomly split into 180 cases for training, 20 cases for validation, and 58 cases for testing. For the training images, only 18 (i.e., 10\%) were used as labeled and the remaining 162 scans were used as unlabeled. For pre-processing, we just normalize each scan to zero mean and unit variance. In the evaluation stage, following existing work~\cite{lin2019deep}, we used the commonly-adopted Dice Similarity Coefficient (\textit{DSC}) and the Average Surface Distance (\textit{ASD}) as segmentation quality evaluation metrics.
\par The framework was implemented in PyTorch~\cite{paszke2019pytorch}, using a node of a cluster with 8 TiTAN 1080TI GPUs. We used the SGD optimizer (weight decay=0.0001, momentum=0.9) with Eq.~\ref{equ:total_loss} for training our method. During the training processing, the poly learning rate strategy was used for learning rate decay, where the initial learning rate 0.1 was multiplied by $(1.0 - \frac{t}{t_{max}}) ^ {\gamma}$ with  $\gamma$ = 0.9 and $t_{max}$ = $60 k$. The batch size was set to 4, and each batch consists of two annotated images and two unannotated images. We randomly cropped $112 \times 112 \times 112$ sub-volumes as the network input and employed data augmentation to enlarge dataset and avoid over-fitting, including random cropping, random flipping and random rotation. The final segmentation results were obtained by using a sliding window strategy.  Following~\cite{yu2019uncertainty}, the $w_{max}$ was set to 0.1 for all experiments. (Details of the NPC dataset and code are presented in supplementary materials.)

\subsection{Evaluation of Our Proposed URPC on the NPC dataset}
\begin{table}[t]
\centering
\footnotesize
\renewcommand\arraystretch{1.0}
\caption{Ablation study of the proposed URPC framework on the NPC MRI dataset, where 18 labeled and 162 unlabeled images were used for training. UR and UM denote the uncertainty rectification term and uncertainty minimization term, respectively.}
\scalebox{0.8}{
\begin{tabular}{lllllll
}
\hline
\multirow{2}{*}{Method} & \multicolumn{2}{c}{GTVnx} & \multicolumn{2}{c}{GTVnd} & \multicolumn{2}{c}{Mean} \\\cline{2-7} 
& \textit{DSC} (\%) &\textit{ASD} (voxel) & \textit{DSC} (\%) &\textit{ASD} (voxel) & \textit{DSC} (\%) &\textit{ASD} (voxel) \\
\hline
Baseline (S = 1)&71.94$\pm$11.60&2.42$\pm$1.65&66.27$\pm$14.62&3.60$\pm$3.12&69.10$\pm$10.15&3.01$\pm$1.76\\
S = 2 & 79.88$\pm$6.91 & 1.79$\pm$1.27 & 72.82$\pm$15.55 & 2.85$\pm$2.54 & 76.35$\pm$9.48 & 2.32$\pm$1.46 \\
S = 3 & 79.09$\pm$5.82 & \textbf{1.76$\pm$0.97} & 75.08$\pm$13.22 & \textbf{2.25$\pm$2.27} & 77.09$\pm$7.85 & \textbf{2.05$\pm$1.24} \\
\textbf{S = 4} & \textbf{80.13$\pm$6.37} & 1.82$\pm$1.30 & \textbf{75.83$\pm$12.93} & 2.65$\pm$2.77 & \textbf{77.98$\pm$8.00} & 2.24$\pm$1.53\\
S = 5 & 79.10$\pm$6.53 & 1.84$\pm$1.14 & 75.73$\pm$13.71 & 2.29$\pm$2.61 & 77.42$\pm$8.27 & 2.06$\pm$1.43\\
\hline
S = 4 + UR & \textbf{80.99$\pm$5.50} & 1.70$\pm$1.12 & 75.22$\pm$13.86 & 3.05$\pm$3.16 & 78.11$\pm$8.06 & 2.38$\pm$1.65 \\
S = 4 + UR + UM & 80.76$\pm$5.72 & \textbf{1.69$\pm$1.06} & \textbf{75.95$\pm$12.74} & \textbf{2.20$\pm$2.07} & \textbf{78.36$\pm$7.66} & \textbf{1.95$\pm$1.18}\\

\hline
\end{tabular}}
\label{tab:abl_tab}
\end{table}
\begin{table}[t]
\centering
\footnotesize
\renewcommand\arraystretch{0.85}
\caption{Comparison between our method and existing methods on the NPC MRI dataset, when using 10\% labeled data. $^*$ denotes $p$-value $<$ 0.05 when comparing the proposed with the others.}
\scalebox{0.85}{
\begin{tabular}{cllllllll
}
\hline
& \multirow{2}{*}{Method} & \multicolumn{2}{c}{GTVnx} & \multicolumn{2}{c}{GTVnd} & \multicolumn{2}{c}{Mean} &
\multirow{2}{*}{\textit{T-T (h)}} \\\cline{3-8} 
& & \textit{DSC} (\%) &\textit{ASD} (voxel) & \textit{DSC} (\%) &\textit{ASD} (voxel) & \textit{DSC} (\%) &\textit{ASD} (voxel)& \\          
\hline
& SL (10\%) &71.94$\pm$11.60$^*$&2.42$\pm$1.65$^*$&66.27$\pm$14.62$^*$&3.60$\pm$3.12$^*$&69.10$\pm$10.15$^*$&3.01$\pm$1.76$^*$&73 \\
&SL (100\%) &83.93$\pm$4.77$^*$&1.35$\pm$0.73$^*$&83.10$\pm$9.05$^*$&1.48$\pm$1.73$^*$&83.51$\pm$5.35$^*$&1.41$\pm$0.94 $^*$&61 \\
& MT~\cite{tarvainen2017mean}&79.80$\pm$6.74$^*$&1.70$\pm$1.17&69.78$\pm$16.34$^*$&2.81$\pm$2.57$^*$&74.79$\pm$9.15$^*$&2.25$\pm$1.40 $^*$&76\\
& ICT~\cite{ijcai2019-504}&80.58$\pm$6.23&1.58$\pm$1.02$^*$&72.62$\pm$13.47$^*$&2.72$\pm$2.61$^*$&76.59$\pm$7.98$^*$&2.15$\pm$1.38 $^*$&78\\
& EM~\cite{vu2019advent}&79.85$\pm$6.32$^*$&1.66$\pm$1.06&69.92$\pm$15.39$^*$&3.14$\pm$2.82$^*$&74.89$\pm$8.85$^*$&2.40$\pm$1.54$^*$&74\\
& UAMT~\cite{yu2019uncertainty}&79.62$\pm$7.16$^*$&1.67$\pm$1.05&71.98$\pm$15.66$^*$&2.55$\pm$2.58$^*$&75.78$\pm$9.67$^*$&2.11$\pm$1.39$^*$ &95\\
& DAN~\cite{zhang2017deep}&80.47$\pm$5.73&\textbf{1.56$\pm$0.81}$^*$&74.62$\pm$12.83$^*$&2.74$\pm$2.62$^*$&77.55$\pm$7.39&2.15$\pm$1.26$^*$ &104\\
& Ours & \textbf{80.76$\pm$5.72} & 1.69$\pm$1.06 & \textbf{75.95$\pm$12.74} & \textbf{2.20$\pm$2.07} & \textbf{78.36$\pm$7.66} & \textbf{1.95$\pm$1.18}&\textbf{74}\\
\hline
\end{tabular}}
\label{tab:sota_10}
\end{table}
\subsubsection{Ablation Study}Firstly, to investigate the impact of different numbers of scales in the pyramid prediction of PPNet, as shown in Fig.~\ref{overview}, we set $S$ of PPNet to 2, 3, 4, and 5, respectively, and UR and UM were not used at this stage. They were compared with the baseline of 3D UNet~\cite{cciccek20163d} without multi-scale predictions and therefore it only learns from labeled data. In contrast, the PPNet learns from both labeled data and unlabeled data. The quantitative results are shown in Tab.~\ref{tab:abl_tab}. It can be found that when $S$ increases from 2 to 4, the performance of the proposed URPC improves gradually. However, we found that $S$  = 5 achieved a lower performance than $S$ = 4. That is because the resolution of $p_4$ is too small to preserve more details. Therefore, we used $S$ = 4 for SSL in the following experiments. Secondly, to measure the contribution of the uncertainty rectifying module, we then turn on the UR term and UM term with $S$ = 4 for training. From the last section of Tab.~\ref{tab:abl_tab}, we can see that both uncertainty rectifying (UR) term and uncertainty minimization (UM) term boost the model performance. What's more, combining all sub-modules into a unified framework results in a better gain where the mean \textit{DSC} and \textit{ASD} were improved by 9.26\% and 1.06 voxels than the baseline, demonstrating their effectiveness for semi-supervised segmentation.
\subsubsection{Comparison with Other Semi-Supervised Methods.} 
\begin{figure}[t]
    \centering
    \includegraphics[width=1.0\textwidth]{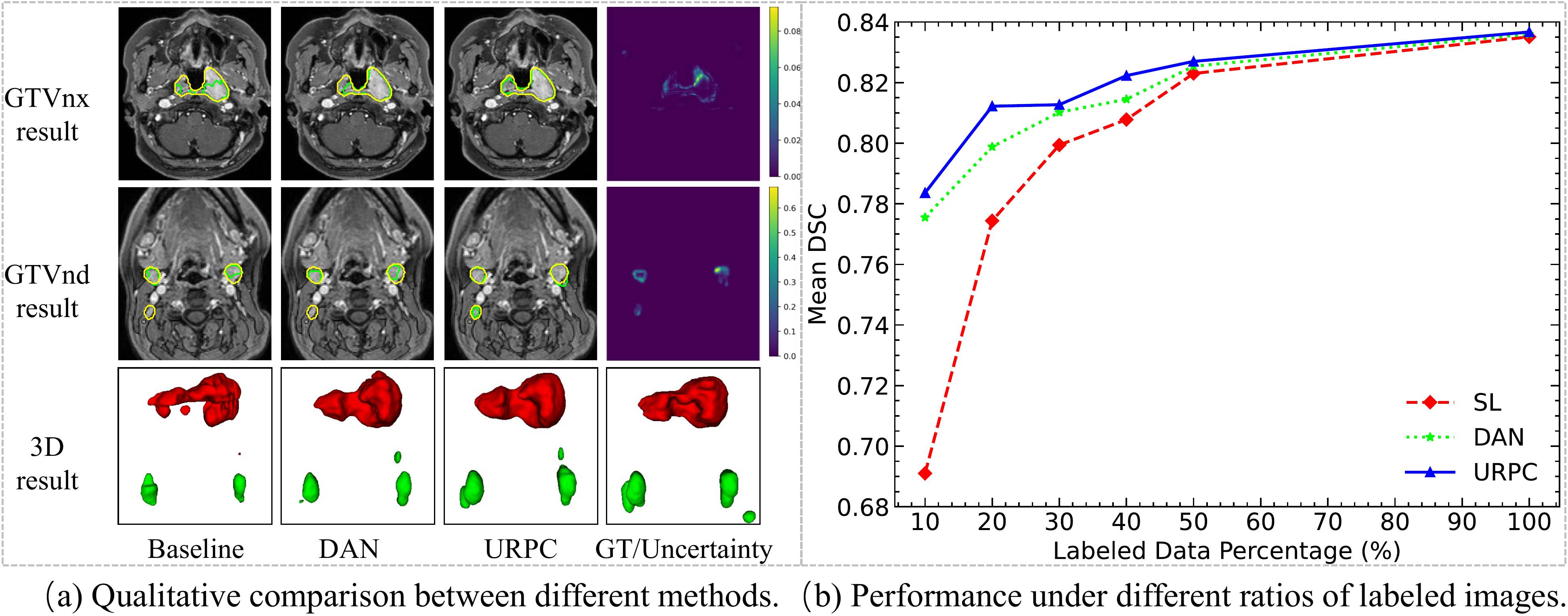}
    \caption{Comparison between different methods}
    \label{fig:vis}
\end{figure}

We compared our method with only using 18 annotated images for supervised learning with 3D UNet , which is denoted as SL (10\%). Similarly, SL (100\%) denote supervised learning with all the training images annotated, which gives the performance upper bound. In addition, we furthercompared our methods with five state-of-the-art semi-supervised segmentation methods, including Mean Teacher (MT)~\cite{tarvainen2017mean}, Interpolation Consistency Training (ICT)~\cite{ijcai2019-504}, Entropy Minimization (EM)~\cite{vu2019advent}, Uncertainty Aware Mean Teacher (UAMT)~\cite{yu2019uncertainty} and Deep Adversarial Network (DAN)~\cite{zhang2017deep}. Note that, for a fair comparison, all these methods were implemented by using 3D UNet~\cite{cciccek20163d} as the backbone and they are online available~\cite{ssl4mis2020}. Tab.~\ref{tab:sota_10} shows the quantitative comparison of these methods. It can be found that compared with SL (10\%), all semi-supervised methods improve the segmentation performance by a large margin, as they can learn from the unannotated data by a regularization loss during the training, and the DAN~\cite{zhang2017deep} achieve the best results among existing methods. Our framework (URPC) achieves better performance than these semi-supervised methods when using 10\% labeled data. These results show that our URPC has the capability to capture the rich information from the unlabeled data in addition to labeled data. What's more, our method is more efficient than existing methods and requires less training time ($T$-$T$) and computational cost, as it just needs to pass an input image once in an iteration. In Fig.~\ref{fig:vis} (a), we visualize some 2D and 3D results of supervised and semi-supervised method when using 10\% labeled data. Compared with supervised learning (SL) baseline and DAN~\cite{zhang2017deep}, our method has a higher overlap ratio with the ground truth and reduces the false negative in both slice level and volume level, especially in GTVnd segmentation. We further visualized the estimated uncertainty ($\mathcal{D}_{0}$ in Eq.~\ref{equ:unc}) in the last column of Fig.~\ref{fig:vis} (a), showing that the uncertain region is mainly distributed near the boundary. We further performed a study on the data utilization efficiency of the URPC. Fig.~\ref{fig:vis} (b) shows the evolution curve of mean \textit{DSC} of GTVnx and GTVnd segmentation obtained by SL, DAN~\cite{zhang2017deep} and URPC when using different numbers of labeled data. It can be found that URPC consistently outperforms SL and DAN~\cite{zhang2017deep}, and when increasing the labeled ratio to 50\%, URPC achieves the mean DSC of 82.74\% which is very close to 83.51\% obtained by SL (100\%). These results demonstrate that the URPC has the capability to utilize the unlabeled data to bring performance gains. More results on 20\% labeled data presented in supplementary materials showed that our method also outperforms these existing methods. 

\section{Conclusion}
In this paper, we proposed a novel efficient semi-supervised learning framework URPC for medical image segmentation. A pyramid prediction network is employed to learn from the unlabeled data by encouraging to produce consistent predictions at multiple scales. An uncertainty rectifying module is designed to improve the stability of learning from unlabeled images and further boost model performance, where the uncertainty estimation can be obtained with a single forward pass efficiently. We applied the proposed method to the segmentation of GTVnx and GTVnd, and the results demonstrated the effectiveness and generalization of URPC and also indicated the promising potential of our proposed approach for further clinical use. In the future, we will evaluate the framework on other segmentation tasks. 
\section{Acknowledgment} This work was supported by the National Natural Science Foundations of China [81771921, 61901084], and also by key research and
development project of Sichuan province, China [20ZDYF2817]. We thank M.D. Mengwan Wu and Yuanyuan Shen from the Sichuan Provincial People's Hospital for the data annotation and refinement.

\section{Appendix}
\subsubsection{Details of NPC dataset and methods' implementations}
\begin{figure}[t]
	\centering
	\includegraphics[width=0.85\linewidth]{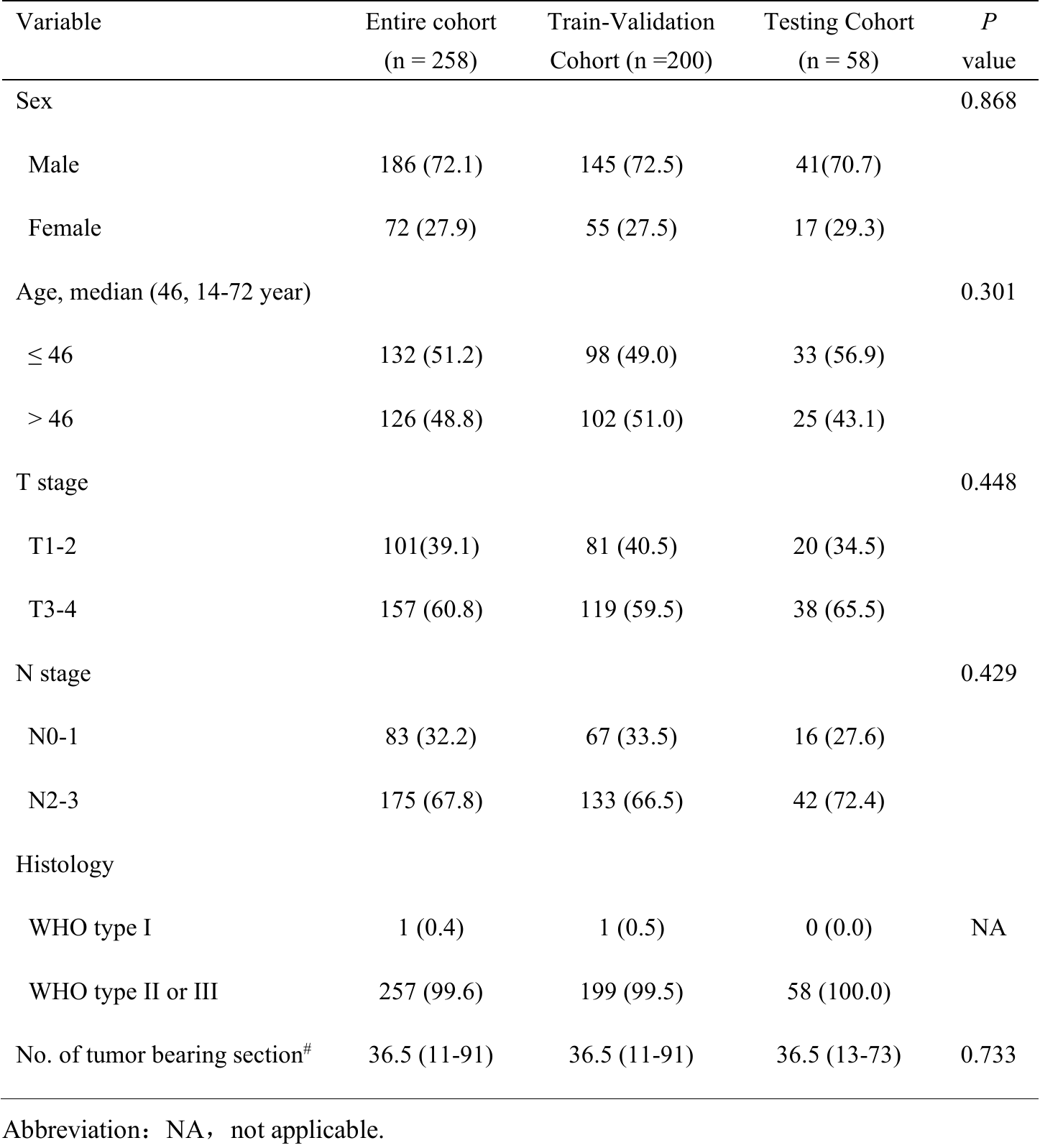}
	\caption{Clinical and Tumor Characteristics of the collected MRI-NPC dataset.}
	\label{fig:details}
\end{figure}

The NPC dataset used in this study was collected from the West China Hospital, Sichuan University, all of them are newly diagnosed NPC patients between January 2010 and December 2016 were reviewed. All MRI images of NPC patients were acquired on 3T Siemens scanners. Due to the large differences in the scanning parameters and section thickness of MRI at different times, in order to obtain uniform and high-quality MRI images, we have made strict inclusion and exclusion criteria. The main inclusion criteria as follows: (1) Histologically confirmed and treatment-naive NPC; (2) MRI examinations were performed with unenhanced T1-, T2-weighted, and contrasted-enhanced T1-weighted sequences; (3) The region of MRI examinations must include the nasopharynx and the neck; (4) The section thickness was between 1-2 mm. The patients will be excluded whose scanned images had a low resolution that affected the delineation of the tumor volume. Finally, 258 patients with NPC were included. The mean resolution of MRI image was 1.23 mm×1.23 mm×1.10 mm, and the mean dimension was 176×286×245. The data set was randomly divided into training set (n = 180), validation set (n = 20) and test set (n = 58). All patients were re-staged according to the eighth edition of the American Joint Committee on Cancer (AJCC)~\cite{edge2010american}. The utilization of clinical data and imaging data was approved by the Ethics Committee of this hospital. The clinical characteristics of the NPC data is presented in Fig.~\ref{fig:details}. The implementation of this work is mainly based on an open-source codebase~\cite{ssl4mis2020}, so it is easy to reproduce the results and compare them with others.

\begin{table}[!htpb]
\centering
\footnotesize
\renewcommand\arraystretch{0.9}
\setlength{\tabcolsep}{0.4mm}{
\begin{tabular}{cllllllll
}
\hline
& \multirow{2}{*}{Method} & \multicolumn{2}{c}{GTVnx} & \multicolumn{2}{c}{GTVnd} & \multicolumn{2}{c}{Mean} &
\multirow{2}{*}{\textit{T-T (h)}} \\\cline{3-8} 
& & \textit{DSC} (\%) &\textit{ASD} (voxel) & \textit{DSC} (\%) &\textit{ASD} (voxel) & \textit{DSC} (\%) &\textit{ASD} (voxel)& \\
\hline
\hline
& SL (20\%) &80.72$\pm$8.28$^*$&1.64$\pm$0.99$^*$&74.16$\pm$15.46$^*$&2.80$\pm$3.10&77.44$\pm$9.99$^*$&2.22$\pm$1.67$^*$ &\textbf{64}\\
&SL (100\%) &83.93$\pm$4.77$^*$&1.35$\pm$0.73$^*$&83.10$\pm$9.05$^*$&1.48$\pm$1.73$^*$&83.51$\pm$5.35$^*$&1.41$\pm$0.94$^*$&61 \\
& MT~\cite{tarvainen2017mean}&81.09$\pm$5.90$^*$&1.54$\pm$0.77&76.30$\pm$13.25$^*$&2.30$\pm$2.36&78.70$\pm$8.03$^*$&1.92$\pm$1.23&72 \\
& ICT~\cite{ijcai2019-504}&81.86$\pm$5.91$^*$&1.53$\pm$0.92&77.42$\pm$12.48$^*$&2.32$\pm$2.57&79.64$\pm$7.43$^*$&1.92$\pm$1.34&69 \\
& EM~\cite{vu2019advent}&82.05$\pm$5.28&1.51$\pm$0.84&77.78$\pm$12.18$^*$&\textbf{2.13$\pm$2.28}&79.92$\pm$7.12$^*$&\textbf{1.82$\pm$1.22} &64\\
& UAMT~\cite{yu2019uncertainty}&81.38$\pm$6.38$^*$&1.60$\pm$0.97$^*$&77.47$\pm$12.55$^*$&2.30$\pm$2.64&79.43$\pm$7.86$^*$&1.95$\pm$1.39 &91\\
& DAN~\cite{zhang2017deep}&81.68$\pm$5.68$^*$&1.53$\pm$0.90&78.09$\pm$12.91$^*$&2.27$\pm$2.37&79.88$\pm$7.41$^*$&1.90$\pm$1.25 &132\\
& Ours & \textbf{82.64$\pm$5.67} & \textbf{1.48$\pm$0.82} & \textbf{79.79$\pm$10.81} & 2.28$\pm$2.66 & \textbf{81.22$\pm$6.43} & 1.88$\pm$1.39&\textbf{66}\\
\hline
\end{tabular}}
\caption{Comparison between our method and existing methods on the NPC MRI dataset, when using 20\% labeled data. $^*$ denotes $p$-value $<$ 0.05 when comparing the proposed with the others.}
\label{tab:sota_20} 
\end{table}

\begin{figure}[htbp]
    \centering
    \includegraphics[width=0.85\textwidth, height=70mm]{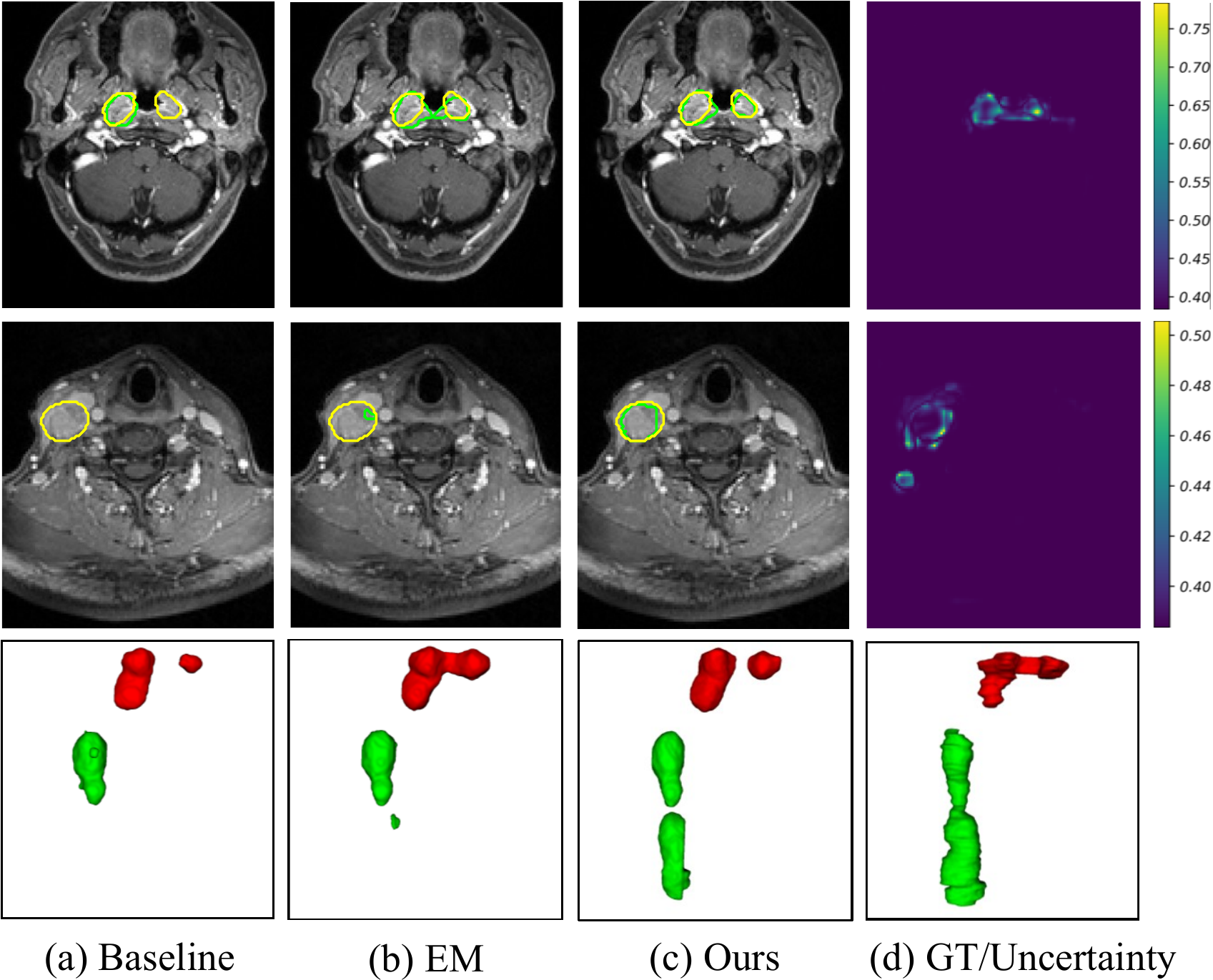}
    \caption{Visualization of the results by different methods and uncertainty map obtained by our method. The results are based on 20\% labeled data and 80\% unlabeled data. Lime and yellow contours denote the prediction and ground truth, respectively. In 3D results, the red and green colors show the GTVnx and GTVnd segmentation, respectively.}
    \label{fig:vis_20}
\end{figure}


\subsubsection{Results on 20\% labeled NPC data.} Following the experiments on 10\% labeled data, we further compared the proposed method with several state-of-the-art semi-supervised segmentation methods using 20\% labeled data. Tab.~\ref{tab:sota_20} shows the quantitative comparison of these methods. It can be found that compared with the SL, all semi-supervised methods improve the segmentation performance by a large margin, as they can learn from the unannotated data by a regularization loss during the training, especially when just used 20\% labeled data, and the EM~\cite{vu2019advent} achieve the best results among existing methods respectively. Our framework (URPC) achieves the best performance over the state-of-the-art semi-supervised methods when using 20\% labeled data. In Fig.~\ref{fig:vis_20}, we visualize some 2D and 3D results of supervised and semi-supervised method when using 20\% labeled data. The first, second and third rows show the GTVnx segmentation, GTVnd segmentation and 3D visualization respectively. Compared with supervised learning and EM~\cite{vu2019advent}, our method has a higher overlap ratio with the ground truth and reduces the false negative in both slice level and volume level, especially in GTVnd segmentation (the second row). We further visualized the estimated uncertainty ($\mathcal{D}_{0}$ in Eq.4) in  Fig.~\ref{fig:vis_20} (d), it can be found that the uncertain region is mainly distributed near the boundary.

\subsubsection{Discussion and future work}
In this work, we found that it is useful for semi-supervised learning to encourage the multi-scale outputs of PPNet to produce consistent predictions. We design a novel pyramid consistency loss to utilize the unlabeled data efficiently. Inspired by existing works~\cite{yu2019uncertainty,wang2019aleatoric,cao2020uncertainty,zheng2020rectifying}, we also use uncertainty maps to select reliable pixels for robust unsupervised learning. Compared with these methods, the advantage of our method as follows: (1) Thanks to the multiple predictions in PPNet, we can measure the uncertainty in a single forward pass by calculating the variance of these predictions, which is much more efficient than the commonly used MC dropout that requires multiple forward passes; (2) Our method automatically selects reliable pixels rather than manually designing a threshold to cut off. In this work, we just apply the URPC for NPC GTV segmentation but it also can be used to segment other lesions and organs as it does not rely on any task-specific knowledge. Recently, some works, such as multi-head/decoder network and grouped convolution-based CNNs~\cite{wang2020uncertainty} can also produce multiple predictions and estimate the model uncertainty in a single forward pass. However, these methods are limited by the computational cost, as the grouped convolution-based CNN and multi-head/decoder network increase the model capacity and require more GPU memory in the same batch size. In addition, they are just designed to deal with interactive refinement and uncertainty estimation respectively and lack of evaluation in semi-supervised learning. 

\bibliographystyle{splncs04}
\bibliography{ref.bib}

\end{document}